\documentclass[letterpaper]{article} 
\usepackage{aaai24}  
\usepackage{times}  
\usepackage{helvet}  
\usepackage{courier}  
\usepackage[hyphens]{url}  
\usepackage{graphicx} 
\usepackage{natbib}  
\usepackage{caption} 

\usepackage{url}            
\usepackage{booktabs}       
\usepackage{amsfonts}       
\usepackage{nicefrac}       
\usepackage{microtype}      
\usepackage{xcolor}         
\usepackage{multirow}
\usepackage{amsmath}
\usepackage{comment}
\usepackage{bm}
\usepackage{lscape}
\usepackage{filecontents}

\usepackage{algorithm}
\usepackage{algorithmic}

\usepackage{newfloat}
\usepackage{listings}
\DeclareCaptionStyle{ruled}{labelfont=normalfont,labelsep=colon,strut=off} 
\floatstyle{ruled}
\newfloat{listing}{tb}{lst}{}
\floatname{listing}{Listing}
\title{Inducing Point Operator Transformer: A Flexible and Scalable Architecture for Solving PDEs}
\author {
    Seungjun Lee,
    Taeil Oh
}
\affiliations {
    Alsemy, South Korea \\
    seungjun.lee@alsemy.com, taeil.oh@alsemy.com
}

\usepackage{bibentry}
\begin{document}

\maketitle

\begin{abstract}
Solving partial differential equations (PDEs) by learning the solution operators has emerged as an attractive alternative to traditional numerical methods. 
However, implementing such architectures presents two main challenges: flexibility in handling irregular and arbitrary input and output formats and scalability to large discretizations. Most existing architectures are limited by their desired structure or infeasible to scale large inputs and outputs. 
To address these issues, we introduce an attention-based model called an inducing-point operator transformer\footnote{Our code is available at \url{https://github.com/7tl7qns7ch/IPOT}.} (IPOT). Inspired by inducing points methods, IPOT is designed to handle any input function and output query while capturing global interactions in a computationally efficient way. 
By detaching the inputs/outputs discretizations from the processor with a smaller latent bottleneck, IPOT offers flexibility in processing arbitrary discretizations and scales linearly with the size of inputs/outputs.
Our experimental results demonstrate that IPOT achieves strong performances with manageable computational complexity on an extensive range of PDE benchmarks and real-world weather forecasting scenarios, compared to state-of-the-art methods. 
\end{abstract}

\section{Introduction}

Partial differential equations (PDEs) are widely used for mathematically modeling physical phenomena by representing pairwise interactions between infinitesimal segments. 
Once formulated, PDEs allow us to analyze and predict the physical system, making them essential tools in various scientific fields \cite{pde}. However, formulating accurate PDEs can be a daunting task without domain expertise where there remain numerous unknown processes for many complex systems. Moreover, traditional numerical methods for solving PDEs can require significant computational costs and may sometimes be intractable.
In recent years, data-driven approaches have emerged as an alternative to the conventional procedures for solving PDEs, since they provide much faster predictions and only require observational data. 
In particular, operator learning, learning mapping between infinite-dimensional function spaces, generalizes well to unseen system inputs with their own flexibility in a discretization-invariant way \cite{deeponet2, operator_com, operator_overview, gno, mgno, fno}.

Observational data in solving PDEs often comes in much more diverse measurement formats, making it challenging to handle. These formats may include sparse and irregular measurements, different input and output domains, and complex geometries due to environmental conditions \cite{irregular1, irregular2, geo_fno, ffno}.
In addition, adaptive remeshing schemes are often deployed when different regions require different resolutions \cite{fem1, fem2}. However, most existing operator learners have their own restrictions. For instance, some require fixed discretization for the input function \cite{deeponet1, deeponet2}, assume that input and output discretizations are the same \cite{operator_com, mgno, fno, mppde}, assume local connectivity \cite{gno, mgno},
or have uniform regular grids \cite{fno, mwt, transformer_galerkin}. They can be limited when the observations have discrepancies between their own settings and given measurements. 
In order to be flexible to handle a variety of discretization schemes, the model should impose minimal assumptions on locality or data structure.

Our approach aims to address the challenges of handling arbitrary input and output formats by developing a mesh-agnostic architecture that treats observations as individual elements without problem-specific modifications. 
As flexible in processing data \cite{perceiver} and efficient in modeling long-range dependencies \cite{transformer_dissection}, Transformer \cite{transformer} can be an appropriate starting point for our approach. A core building block of the Transformers, the attention mechanism corresponds to and even outperforms the kernel integral operation of the existing operator learners due to its nature of capturing long-range interactions \cite{transformer_galerkin, htnet, transformermeet}. 
However, their quadratic growth in computational complexity with input size can make it impractical for real-world applications, high-fidelity modeling, or long-term forecasting without imposing problem-specific locality.

To address the issue, we took inspiration from inducing-point methods which aim to reduce the effective number of input data points for computational efficiency \cite{ip0, ip1, ip2, mgno}, and from the extension of the methods to Transformer by cross-attention with a small number of learnable queries \cite{set_transformer, sensory_neuron, perceiver, perceiver_io, ip_transformer}.
In this paper, we introduce a fully attention-based model called an inducing-point operator transformer (IPOT) to capture long-range interactions and provide flexibility for arbitrary input and output formats with feasible computational complexity.
IPOT consists of an encoder-processor-decoder \cite{epd1, epd2}, where the encoder compresses the input function into a smaller fixed number of latent bottlenecks inspired by inducing-point methods, the processor operates on the latent features, and the decoder produces solutions at any output queries from the latent features.
This architecture achieves scalability by separating input and output discretizations from the latent processor. It allows the architecture to avoid quadratic complexity and decouples the depth of the processor from the size of the inputs or outputs. 
This approach can be used for real-world applications with large, complex systems or long-term forecasting tasks, making it feasible and practical to use. 

Finally, we conducted several experiments on the extensive PDE benchmarks and a real-world dataset. Compared to state-of-the-art methods, IPOT achieves competitive performances with feasible computational complexity. It can handle uniform regular grids, irregular grids, real-world weather forecasting, and their variants with arbitrary discretization formats. 

\section{Related works}
\textbf{Operator learning.}
\label{related_works_operator_learning}
Based on a pioneer work \cite{onet}, deep operator networks (DeepONets) were presented to extend the architectures for operator learning with modern deep networks \cite{deeponet1, deeponet2}. DeepONets consists of a branch network and a trunk network and can be queried out at any coordinate from the trunk network. However, they require the fixed discretization of the input functions from the branch network \cite{operator_com}.
Another promising framework is neural operators, which consist of several integral operators with parameterized kernels to map between infinite-dimensional functions \cite{operator_overview, gno, mgno, fno}. 
Message passings on graphs \cite{gno, mgno, epd2}, convolutions in the Fourier domain \cite{fno}, or wavelets domain \cite{mwt, wno} are used to approximate the kernel integral operations. However, the implemented architectures typically require having the same sampling points for input and output \cite{operator_com, operator_overview, mgno, fno}.
In addition, graph-based methods do not converge when problems become complex, and convolutional in the spectral domains methods are limited to the uniform regular grids due to the usage of FFT \cite{fno}. To handle irregular grids, \cite{geo_fno, ffno} apply adaptive coordinate maps before and after FNO to make it extend to irregular meshes. 
Recently, Transformers have been recognized as flexible and accurate operator learners \cite{transformer_galerkin}. However, their quadratic scaling poses a significant challenge to their practical use. \cite{transformer_galerkin} removes the softmax normalization and introduces Galerkin-type attention to achieve linear scaling. Although efficient transformers \cite{htnet, oformer, gnot} have been proposed with preserving permutation symmetries \cite{mino}, the demand for flexibility and scalability is still not met for practical use in the real world. 
\\
\textbf{Inducing-point methods.}
Inducing-point methods have been commonly used to approximate the Gaussian process through $m$ inducing points, where they involve replacing the entire dataset with smaller subsets that are representative of the data. 
These inducing points methods have been widely used in regression \cite{id_regression}, kernel machines \cite{id_kernel}, and matrix factorizations \cite{id_matrix}. A similar idea is also employed in the existing neural operator \cite{mgno}, which is based on graph networks and uses sub-sampled nodes from the original graph as inducing points to reduce the computational complexity of the previous graph neural operator \cite{gno}. However, this approach did not converge for complex problems \cite{fno}. 
Recently, the methods have been extended to Transformers by incorporating cross-attention with a reduced number of learnable query vectors \cite{set_transformer, sensory_neuron, perceiver, perceiver_io, ip_transformer}.

\section{Preliminaries}
\subsection{Neural operators}
Let us consider a set of $N$ input-output pairs, denoted by ${(a_i, u_i)}_{i=1}^N$, where $a_i = a_i|_{X}$ and $u_i = u_i|_{Y}$ are finite discretizations of the input function $a_i \in \mathcal{A}$ at the set of input points $X = \{ x_1, \ldots, x_{n_x} \}$ and output function $u_i \in \mathcal{U}$ at the set of output points $Y = \{ y_1, \ldots, y_{n_y} \}$ with the number of discretized points $n_x$ and $n_y$, respectively. Here, $\mathcal{A}$ and $\mathcal{U}$ are input and output function spaces, respectively. The objective of operator learning is to minimize the empirical loss $E_{a \sim \mu} \left[ \mathcal{L} \left( \mathcal{G}_{\theta} (a), u \right) \right] \approx \frac{1}{N} \sum_{i=1}^{N} \| u_i - \mathcal{G}_{\theta} (a_i) \|^2$ to learn a mapping $\mathcal{G}_{\theta}: \mathcal{A} \rightarrow \mathcal{U}$.
The architectures of the neural operator \cite{operator_overview, fno}, usually consist of three components, namely lifting, iterative updates, and projection, which correspond to the encoder-processor-decoder, respectively. 
\begin{equation}
u =  \mathcal{G}_\theta (a) = (\mathcal{Q} \circ \mathcal{G}_{L-1} \circ \dots \circ \mathcal{G}_1 \circ \mathcal{P})(a)
\label{operator_learning_NO}
\end{equation}
where the lifting (encoder) $\mathcal{P} = \mathcal{G}_{enc}$ and projection (decoder) $\mathcal{Q} = \mathcal{G}_{dec}$ are local transformations that map input features to target dimensional features, implemented by point-wise feed-forward neural networks. The iterative updates $\mathcal{G}_l: v_l \mapsto v_{l + 1}, l \in [0, L-1]$ are global transformations that capture the interactions between the elements, implemented by a sequence of following transformations,
\begin{equation}
\begin{aligned}
\label{iterative_update}
v_{l+1}(x) = \sigma \biggl( \mathcal{W}_l v_l(x) + \left[ \mathcal{K}_l (v_l) \right] (x) \biggr), \quad x \in \Omega,
\end{aligned}
\end{equation}
where $\sigma$ are nonlinear functions, $\mathcal{W}_l$ are point-wise linear transformations, $\mathcal{K}_l$ are kernel integral operations on $v_l(x)$. The existing implementations of the neural operators typically use the same discretizations for the input and output \cite{operator_com, mgno, fno, mppde}, which leads to the predicted outputs only being queried at the input meshes. 
Our core idea is to replace the encoder $\mathcal{G}_{enc}$ and decoder $\mathcal{G}_{dec}$ to accommodate arbitrary size and structure of the input functions and output queries.

\subsection{Kernel integral operation and attention mechanism}
The kernel integral operations are generally implemented by integration of input values weighted by kernel values $\kappa$ representing the pairwise interactions between the elements on input domain $x\in\Omega_x$ and output domain $y\in\Omega_y$,
\begin{equation}
\begin{aligned}
\label{integral_operator}
\left[ \mathcal{K} (v) \right](y) = \int_{\Omega_x} \kappa(y, x) v(x) dx,  \quad (x, y)\in \Omega_x \times \Omega_y,
\end{aligned}
\end{equation}
where the kernel $\kappa$ is defined on $\Omega_y \times \Omega_x$. The transform $\mathcal{K}$ can be interpreted as mapping a function $v(x)$ defined on a domain $x \in \Omega_x$ to the function $[\mathcal{K} (v)](y)$ defined on a domain $y \in \Omega_y$. 
Recently, it has been proved that the kernel integral operation can be successfully approximated by the attention mechanism of Transformers both theoretically and empirically \cite{operator_overview, transformer_galerkin, operator_adaptive, operator_fourcast}. Intuitively, let input vectors $X\in \mathbb{R}^{n_x \times d_x}$ and query vectors $Y\in \mathbb{R}^{n_y \times d_y}$, then the attention can be expressed as
\begin{equation}
\begin{aligned}
\label{attention_mechanism}
\textsl{Attn}(Y, X, X) = \sigma (QK^T)V 
\approx \int_{\Omega_x} (q(Y) \cdot k(x)) v(x) dx,
\end{aligned}
\end{equation}
where $Q = YW^q \in \mathbb{R}^{n_y \times d}$, $K = XW^k \in \mathbb{R}^{n_x \times d}$, $V = XW^v \in \mathbb{R}^{n_x \times d}$, and $\sigma$ are the query, key, value matrices, and softmax function, respectively. $W^q$, $W^k$, and $W^v$ are learnable weight matrices that operate in a pointwise way which makes the attention module not depend on the input and output discretizations. A brief derivation of Equation~\ref{attention_mechanism} can be found in the Appendix.
The weighted sum of $V$ with the attention matrix $\sigma (QK^T)$ can be interpreted as the kernel integral operation in which the parameterized kernel is approximated by the attention matrix \cite{transformer_dissection, transformer_galerkin, Nystrom, transformer_performers}.
This attention is also known as cross-attention, where the input vectors are projected to query embedding space by the attention, \textsl{Attn}$(Y, X, X)$. 
However, the computational complexity of the mechanism is proportional to $O(n_x n_y d)$, and it can cause quadratic complexity $O(n^2 d)$ for large $n_x$ and $n_y$ ($n_x, n_y \approx n$).

\section{Approach}
The goal of this work is to develop a mesh-agnostic architecture that can handle any input and output discretizations with reduced computational complexity. 
We allow for different, irregular, and arbitrary input and output formats.
\\
\textbf{Handling arbitrary input function and output queries.} The output solution evaluated at output query $y$ can be expressed as $u(y) = \left[ \mathcal{G}_\theta (a) \right](y)$ which can be viewed as the operating $\mathcal{G}: \mathcal{A} \times Y \rightarrow \mathcal{U}$ an input function $a \in \mathcal{A}$ at the corresponding output query $y\in Y$. We treat the representations of discretized input $a = a|_X \in \mathbb{R}^{n_x \times d_a}$, output $u = u|_{Y} \in \mathbb{R}^{n_y \times d_u}$ and corresponding output queries $Y \in \mathbb{R}^{n_y \times d_y}$ as sets of individual elements. In practice, they are represented by flattened arrays, without using structured bias to avoid our model bias toward specific data structures and flexibly process any discretization formats. The significant modifications from the existing neural operators (Equation~\ref{operator_learning_NO}) are mostly in the encoder and decoder for detaching the dependence of input and output discretizations by
\begin{equation}
\begin{aligned}
\label{overall_epd}
 Z_0 = \mathcal{G}_{enc} (a), \quad  Z_{l+1} = \mathcal{G}_{l} (Z_l), \quad \tilde{u} = \mathcal{G}_{dec} (Y, Z_L).
\end{aligned}
\end{equation}
where the encoder projects the input function into latent space, the sequence of the processing is operated in latent space, and then the decoder predicts the output solutions from the latent representations at the corresponding output queries.
\\
\textbf{Positional encoding}. In order to compensate for the positional information at each function value, we follow a common way of existing neural operators \cite{operator_overview, fno}, which involves concatenating the position coordinates with the corresponding function values to form the input representation, $a= \{ (x_1, a(x_1)), ..., (x_{n_x}, a(x_{n_x}))\}$. Additionally, instead of using raw position coordinates for both input $X= \{x_1, ..., x_{n_x} \}$ and outputs $Y = \{ y_1, ..., y_{n_y} \}$, we concatenate additional Fourier embeddings for the position coordinates. This technique, which is commonly used to enrich positional information in neural networks \cite{transformer, nerf, fourier_features, siren} involves exploiting sine and cosine functions with frequencies spanning from minimum to maximum frequencies, thereby covering the sampling rates for the corresponding dimensions. 

\begin{figure*}[t]
\centering
\includegraphics[scale=0.65]{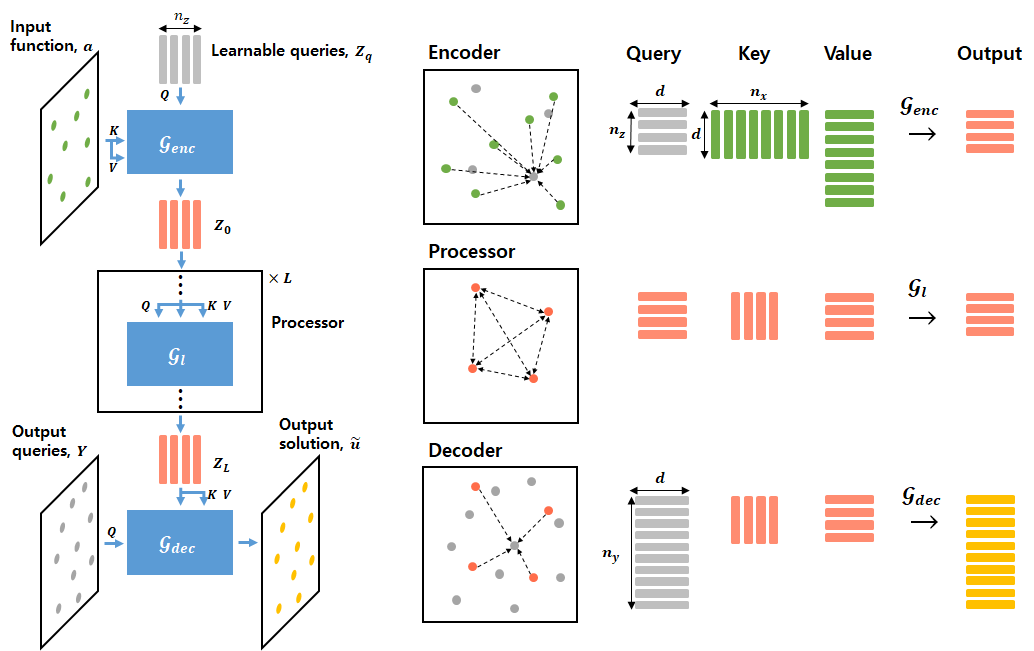}
\caption{Inducing-point operator transformer (IPOT) uses a smaller number of inducing points, enabling it to flexibly handle any discretization formats of input and output, and significantly reduce the computational costs. IPOT encodes input function discretized on
$X= \{x_1, ..., x_{n_x} \}$ to a fixed smaller size $n_z$ of learnable query vectors, and decoding them to output
discretized on $Y= \{y_1, ..., y_{n_y} \}$. The number of size is varied as $n_x$ (arbitrary) → $n_z$ (fixed) → $n_y$ (arbitrary).}
\label{IPOT}
\vspace{-2mm}
\end{figure*}

\subsection{Inducing point operator transformer (IPOT)}
We build our model with an attention-based architecture consisting of an encoder-processor-decoder, called an inducing point operator transformer (IPOT) to reduce the computational complexity of attention mechanisms (Figure~\ref{IPOT}).
The key feature of IPOT is the use of a reduced number of inducing points ($n_z \ll n_x, n_y$) \cite{set_transformer, perceiver, ip_transformer}. This is typically achieved by employing $n_z$ learnable query vectors into the encoder, allowing most of the attention mechanisms to be computed in the smaller latent space instead of the larger observational space. This results in a significant reduction in computational complexity. 
The encoder encodes the input function $a$ to the fixed smaller number of latent feature vectors (discretization number: ${n_x} \mapsto {n_z}$),
the processor processes the pairwise interactions between elements of the latent features vectors (discretization number: ${n_z} \mapsto {n_z}$), and the decoder decodes the latent features to output solutions at a set of output queries $Y$ (discretization number: ${n_z} \mapsto {n_y}$).
\\
\textbf{Attention blocks.} The exact form of the attention blocks \textsl{Attention}$(Y, X, X)$ used in the following sections are described in Appendix.
The nonlinearity $\sigma$, pointwise linear transformations $\mathcal{W}_l$, and kernel integral operations $\mathcal{K}_l$ in Equation~\ref{iterative_update} are approximated by feed-forward neural networks, residual connections, and the attention modules, which are common block forms of Transformer-like architectures \cite{transformer}. In addition, layer normalization \cite{layernorm} is used to normalize the query, key, and values inputs, and multi-headed extensions of the attention blocks can be optionally employed to improve the model's performance. 
\\
\textbf{Encoder.}
We use a cross-attention block as the encoder to encode inputs $a$ with the size $n_x$ to a smaller fixed number $n_z$ of learnable query vectors (inducing points) $Z_{q} \in \mathbb{R}^{n_z \times d_z}$ (typically $n_z \ll n_x$). Then the result of the block is
\begin{equation}
\begin{aligned}
\label{encoder}
 Z_0 = \mathcal{G}_{enc} (a) = \textsl{Attention}(Z_{q}, a, a) \in \mathbb{R}^{n_z \times d_z},
\end{aligned}
\end{equation}
where each learnable query vector of $Z_{q}$ is randomly initialized by a normal distribution and learned along the architectures.
The encoder maps the input domain to a latent domain consisting of $n_z$ inducing points based on the correlations between input discretizations and inducing points. The computational complexity of the encoder is proportional to $O(n_x n_z d)$ which scales linearly with the input size $n_x$. Furthermore, the use of inducing points reduces the computational complexity of the subsequent attention blocks.
\\
\textbf{Processor.}
We use a series of self-attention blocks as the processor each of which takes $Z_l \in \mathbb{R}^{n_z \times d_z}$ as the input of the query, key, and value components. Then the output of each self-attention block for $l \in [0, L - 1]$ is
\begin{equation}
\begin{aligned}
\label{processor}
Z_{l + 1} = \mathcal{G}_{l} (Z_l) = \textsl{Attention}(Z_l, Z_l, Z_l) \in \mathbb{R}^{n_z \times d_z},
\end{aligned}
\end{equation} 
which captures global interactions of inducing points in the latent space. 
Since the processor is decoupled from the input and output discretizations, IPOT is not only applicable to any input and output discretization formats but also significantly reduces computational costs.
Processing in the latent space rather than in observational space reduces the costs in the processor to $O(L n_z^2 d)$ from $O(L n_x^2 d)$ for the original Transformers. This decouples the depth of the processor $L$ from the size of input $n_x$ or output $n_y$, allowing the construction of deep architectures or long-term forecasting models with large $L$.
\\
\textbf{Decoder.}
We use a cross-attention block as the decoder to decode the latent vectors from the processor $Z_{L} \in \mathbb{R}^{n_z \times d_z}$ at output queries $Y \in \mathbb{R}^{n_y \times d}$. Then, the output solutions predicted by IPOT at the corresponding output queries are
\begin{equation}
\begin{aligned}
\label{decoder}
\tilde{u} = \mathcal{G}_{dec} (Y, Z_L) = \textsl{Attention}(Y, Z_L, Z_L) \in \mathbb{R}^{n_y \times d_u}.
\end{aligned}
\end{equation}
The decoder maps the latent domain to the output domain based on the correlations between $n_z$ inducing points and output queries $Y$. 
Since the result from the processor is independent of the discretization format of input function $a$, the entire model is also applicable to arbitrary input discretization and can be queried out at arbitrary output queries.
In addition, the computational cost of the decoder is proportional to $O(n_z n_y d)$ which also scales linearly with the output size $n_y$. 

\begin{figure*}[t]
\centering
\includegraphics[scale=0.65]{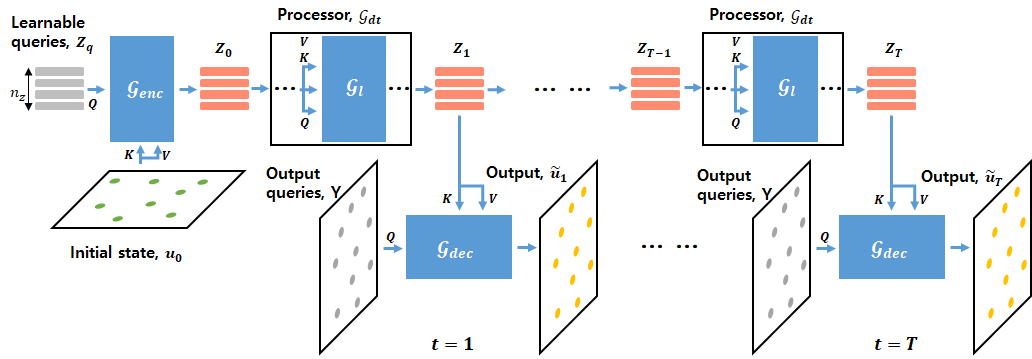}
\caption{The state of the system at time $T$ is denoted as $u_T = ( \mathcal{G}_{dec} \circ [\mathcal{G}_{dt} \circ \cdots \circ \mathcal{G}_{dt}] \circ \mathcal{G}_{enc} ) (u_0)$, where $\mathcal{G}_{enc}$ is an encoder that maps the initial state $u_0$ to the latent state, $\mathcal{G}_{dec}$ is a decoder that maps the latent state back to the observational state, and $\mathcal{G}_{dt}$ is a processor that steps forward in time by $dt$ implemented by a series of self-attention blocks on latent states. }
\label{IPOT_time}
\vspace{-2mm}
\end{figure*}

\subsection{Time-stepping through latent space}
We model the time-dependent PDEs as an autoregressive process. The state of the system at time $t+dt$ is described as $u_{t+dt} = (\mathcal{G}_{dec} \circ \mathcal{G}_{dt} \circ \mathcal{G}_{enc}) (u_t)$, where $\mathcal{G}_{enc}$ is an encoder that maps the current observational state to the latent state, $\mathcal{G}_{dec}$ is a decoder that maps the latent state back to the observational state, and $\mathcal{G}_{dt}$ is a processor that steps forward in time by $dt$ implemented by a series of self-attention blocks on the latent states. We assume that the processor $\mathcal{G}_{dt}$ is independent of $t$ \cite{epd1, epd2, operator_markov, oformer}. 
When setting the time step as $dt = 1$, the predicted trajectory of the system is obtained by the following recurrent relation at each time $t \in [0, T - 1]$
\begin{equation}
\begin{aligned}
\label{induction}
 Z_0 = \mathcal{G}_{enc} (u_0), \quad Z_{t+1} = \mathcal{G}_{dt} (Z_t), \quad \tilde{u}_{t + 1} = \mathcal{G}_{dec} (Y, Z_{t + 1}).
\end{aligned}
\end{equation}
where $u_0$ is the initial state, $Y$ are output queries, and identical processor $\mathcal{G}_{dt}$ is applied at each time step. Throughout the entire trajectory, by encoding the initial state into the latent space, we can significantly reduce the computational costs of subsequent processing compared to processing in the observational space. 

\subsection{Computational complexity}
The overall computational complexity of IPOT is proportional to $O(n_x n_z d + L n_z^2 d + n_z n_y d)$ which scales linearly to $n_x$ and $n_y$, and the depth of architecture $L$ is decoupled from the input and output size. When $n_x, n_y \approx n$ and $L \gg 1$, the complexity becomes $O(Ln_z^2 d + 2n n_z d) \approx O(L n_z^2 d)$, which is significantly lower than that of standard Transformer $O(L n^2 d)$ \cite{transformer}, which has quadratic scaling and is coupled with $L$. Existing operator learners including Fourier neural operator $O(L n\log{n} \,d)$ \cite{fno} or linear Transformers $O(Ln d^2)$ \cite{transformer_galerkin, oformer} have sub-quadratic scaling with the size $n$, but they are coupled with the depth $L$. The decoupling of the depth $L$ from the size $n$ makes it practical to construct deep architectures or apply them to long-term forecasting that requires a large $L$. 

\section{Experiments}
\label{experiments}
We conduct experiments on several benchmark datasets, including PDEs on regular grids \cite{fno}, irregular grids \cite{geo_fno, ffno, dino}, and real-world data from the ERA5 reanalysis dataset \cite{era5} to investigate the flexibility and scalability of our model through various downstream tasks. 
Details of the equations and the problems are described in Appendix.
Results for experiments already discussed on baselines were obtained from the related literature, and results for the extended tasks that have not been discussed before have been reproduced from their original codes. We provide some illustrations of the predictions of IPOT for the benchmarks on regular and irregular grids (Figure~\ref{figure3}) and for long-term dynamics (Figure~\ref{figure4}). 
The implementation details and additional results can be found in the Appendix.
\\
\textbf{Baselines.}
We took several representative architectures as baselines including deep operator network (DeepONet) \cite{deeponet1}, the mesh-based learner with graph neural networks (Meshgraphnet) \cite{epd2}, Fourier neural operator (FNO) \cite{fno}, Factorized-Fourier neural operator (FFNO) \cite{ffno}, and operator Transformer (OFormer) \cite{oformer}. 
\\
\textbf{Evaluation metric.}
We use relative $L_2$ error for the objective functions and evaluation metrics for test errors, where we follow the convention of related literature \cite{operator_overview, fno}.
When $N$ is the dataset size of input-output pairs $(a_i, u_i)_{i=1}^N$, the relative $L_2$ error is defined as 
\begin{equation}
\begin{aligned}
E_{a \sim \mu} [ \mathcal{L} (\mathcal{G}_\theta (a), u) ] = \frac{1}{N} \sum_{i=1}^N \frac{ \| u - \mathcal{G}_\theta (a) \|_2 }{ \| u_i \|_2 }.
\end{aligned}
\end{equation}

\subsection{Problems of PDEs solved on regular grids}
We conducted experiments on benchmark problems, where the PDEs for Darcy flow, and Navier-Stokes equation were solved on regular grids from \cite{fno}. 
Shown in Table~\ref{performance_table}, IPOT consistently demonstrates competitive or strong performances with efficient computational resources, particularly in the case of Navier-Stokes benchmarks that involve higher complexity and time-dependence, resulting in larger $n$ and requiring models with long sequences $L$ of processors or iterative updates. 
These achievements are obtained without using any problem-specific layers such as pooling layers or convolutional filters, which are only compatible the regular grid structures. Consequently, IPOT offers greater flexibility in handling arbitrary inputs and output formats beyond the regular grids.  

\subsection{Problems of PDEs solved on irregular grids}
\begin{figure*}[t]
\centering
\vspace{-3mm}
\includegraphics[scale=0.49]{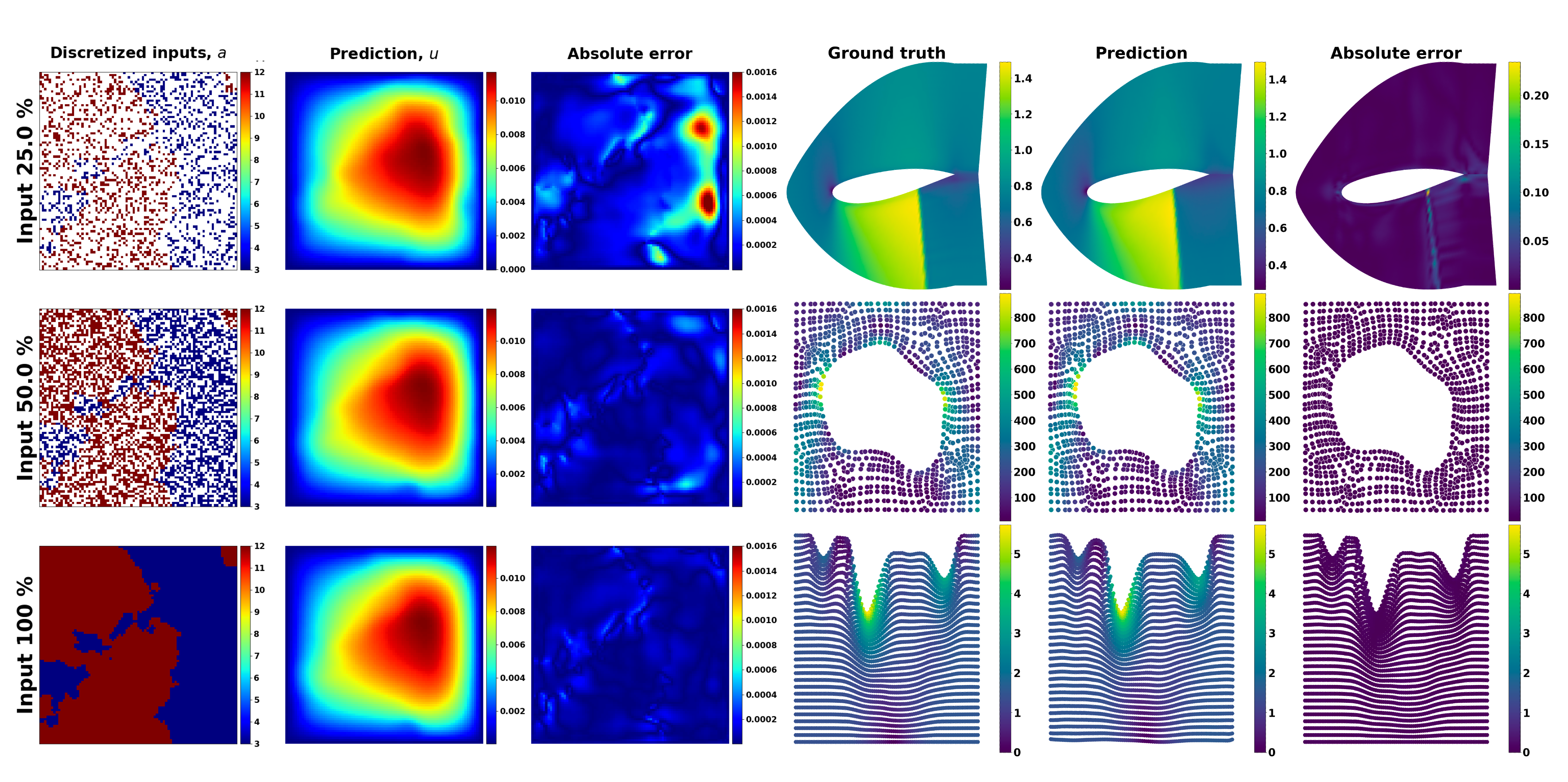}
\caption{Predictions of IPOT on the problems of Darcy flow (left), airfoil (top right), elasticity (middle right), and plasticity (bottom right). In the case of Darcy flow, the partially observed inputs are randomly subsampled with the ratios of $\{100, 50, 25\}\%$.}
\label{figure3}
\end{figure*}

In addition, we conducted experiments on various problems, where the PDEs were solved on non-uniform and irregular grids. These problems include predicting flows around complex geometries (airfoil), estimating stresses on irregularly sampled point clouds (elasticity), and predicting displacements given initial boundary conditions for plastic forging problem (plasticity) as described in \cite{ffno}. 
The experimental results presented in Table~\ref{performance_table} also demonstrate that our IPOT model demonstrates competitive or stronger performances with efficient computational resources, as illustrated in Figure~\ref{figure3}.

\begin{table}[t]
\scriptsize
\caption{Performance and efficiency comparisons with baselines across various datasets. The efficiencies are compared in terms of the number of parameters, time spent per epoch (seconds), and CUDA memory consumption (GB). The missing entries occur when the methods are not able to handle the datasets or when encountering convergence issues.}
\label{performance_table}
\centering
\begin{tabular}{c|c|cccc}
\specialrule{.15em}{.075em}{.075em}
Dataset & Model & Params & Runtime & Memory & Error\\
\specialrule{.15em}{.075em}{.075em}
\multirow{6}{*}{Darcy Flow} & DeepONet     & 0.42M & 2.73 & 2.40 & 4.61e-2 \\
                            & Meshgraphnet & \underline{0.21M} & 9.51 & 5.57 & 9.67e--2 \\
                            & FNO          & 1.19M & \textbf{1.88} & \underline{1.96} & \underline{1.09e--2} \\ 
                            & FFNO         & 0.41M & 3.36 & 1.99 & \textbf{7.70e--3} \\
                            & OFormer      & 1.28M & 3.63 & 5.71 & 1.26e--2 \\
                            & IPOT (ours)  & \textbf{0.15M} & \underline{2.70} & \textbf{1.82} & 1.73e--2 \\
\specialrule{.15em}{.075em}{.075em}
\multirow{6}{*}{Navier Stokes} & DeepONet  &   -   &  -     &  -   &    -     \\
                            & Meshgraphnet & 0.29M & 137.17 & 6.15 & 1.29e--1 \\
                            & FNO          & 0.93M & \underline{53.73}  & \underline{3.09} & 1.28e--2 \\
                            & FFNO         & \underline{0.27M} & 53.82  & 3.40 & 1.32e--2 \\
                            & OFormer      & 1.85M & 70.15  & 9.90  & \underline{1.04e--2} \\
                            & IPOT (ours)  & \textbf{0.12M} & \textbf{21.05} & \textbf{2.08} & \textbf{8.85e--3} \\
\specialrule{.15em}{.075em}{.075em}
\multirow{6}{*}{Airfoil}    & DeepONet     & 0.42M & 3.13 & 2.75  & 3.85e--2 \\
                            & Meshgraphnet & \underline{0.22M} & 10.92 & 6.38 & 5.57e--2 \\
                            & FNO          & 1.19M & \underline{2.63} & \underline{2.73} & 4.21e--2 \\
                            & FFNO         & 0.41M & 4.71 & 2.78 & \textbf{7.80e--3} \\
                            & OFormer      & 1.28M & 4.17  & 7.97 & 1.83e--2 \\
                            & IPOT (ours)  & \textbf{0.12M} & \textbf{2.15} & \textbf{2.10} & \underline{8.79e--3} \\
\specialrule{.15em}{.075em}{.075em}
\multirow{6}{*}{Elasticity} & DeepONet     & 1.03M & 3.72 & \underline{1.18} & 9.65e--2 \\
                            & Meshgraphnet & \underline{0.46M} & 7.36 & 4.04 & 4.18e--2 \\
                            & FNO          & 0.57M & \textbf{1.04} & 1.68 & 5.08e--2 \\
                            & FFNO         & 0.55M & 2.42 & 2.08 & 2.63e--2 \\
                            & OFormer      & 2.56M & 5.58 & 2.98 & \underline{1.83e--2} \\
                            & IPOT (ours)  & \textbf{0.12M} & \underline{1.99} & \textbf{1.13} & \textbf{1.56e--2} \\
\specialrule{.15em}{.075em}{.075em}
\multirow{6}{*}{Plasticity} & DeepONet     &   -   &   -  &  -   &     -    \\
                            & Meshgraphnet &   -   &   -   &   -   &  -  \\
                            & FNO          & 1.85M & \underline{10.40} & 16.81 & 5.08e--2 \\
                            & FFNO         & 0.57M & 66.47 & 16.86 & \underline{4.70e--3} \\
                            & OFormer      & \underline{0.49M} & 28.43 & \underline{14.11} & 1.83e--2 \\
                            & IPOT         & \textbf{0.13M} & \textbf{10.14} & \textbf{5.35} & \textbf{3.25e--3} \\
\specialrule{.15em}{.075em}{.075em}
\multirow{6}{*}{ERA5}   & DeepONet     &   -   &   -   &   -   &    -     \\
                        & Meshgraphnet & 2.07M & 51.75 & 18.45 & 7.16e--2 \\
                        & FNO          & 2.37M & \textbf{9.23} & 13.04  & 1.21e--2 \\
                        & FFNO         & \underline{1.12M} & 14.39 & 17.06 & \underline{7.25e--3} \\
                        & OFormer      & 1.85M & 71.18 & \underline{10.90} & 1.15e--2 \\
                        & IPOT (ours)  & \textbf{0.51M} & \underline{9.83} & \textbf{10.58} & \textbf{6.64e--3} \\
\specialrule{.15em}{.075em}{.075em}
\end{tabular}
\vspace{-3mm}
\end{table}

\subsection{Application to real-world data}
\begin{figure*}[t]
\centering
\includegraphics[scale=0.40]{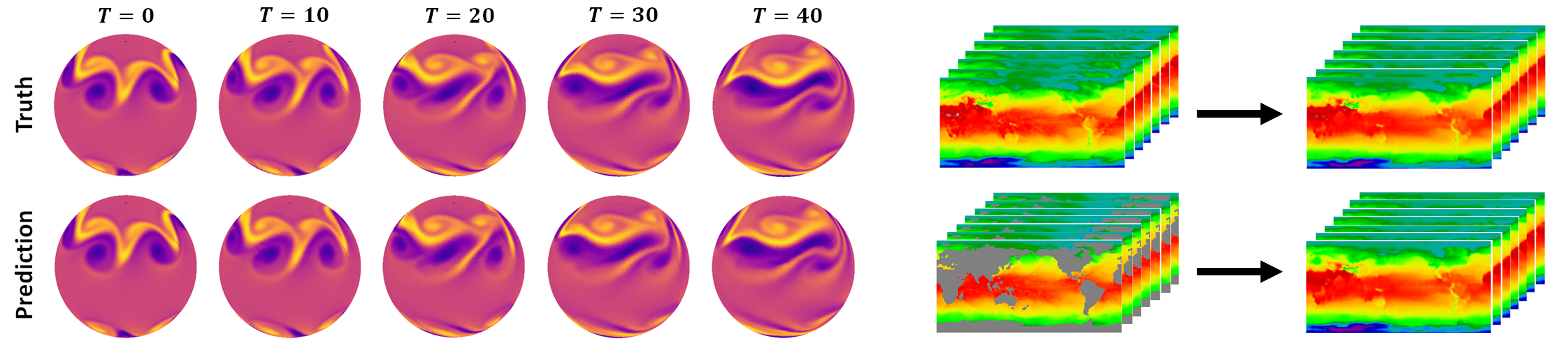}
\caption{Long-term predictions of IPOT on spherical manifolds for the shallow-water equations (left), and on real-world weather forecasting when the inputs are spatially fully given (top right), and partially given (bottom right).}
\label{figure4}
\vspace{-3mm}
\end{figure*}

Unlike the previous problems, we conducted experiments on a subset of the ERA5 reanalysis dataset from the European Centre for Medium-Range Weather Forecasts (ECMWF) \cite{era5}, where the governing PDEs are unknown.
While the ERA5 database includes extensive hourly and monthly measurements for a number of parameters at various pressure levels, our focus is specifically on predicting the daily temperature at 2m from the surface $T_{2m}$. 
IPOT also achieves superior accuracy compared to the baselines while maintaining comparable both time and memory costs. 
Using the decoupled $n_z$ inducing points from the observational space, our approach mitigates the computational burden associated with large $n$ and $L$, making it beneficial for complex and long-term forecasting tasks. 
\\
\textbf{Generalization ability on discretizations.}
Furthermore, we explore the model's generalization ability to different resolutions and to make predictions from partially observed inputs.   
The evaluation of different resolutions is motivated by the scenario where observations are collected at varying resolutions. The evaluation with masked inputs is motivated by situations when observations are only available for the sea surface while data from land areas are unavailable as shown in the bottom right of Figure~\ref{figure4}.
We employ bilinear interpolation methods to obtain interpolated input values corresponding to land coordinates and combine them with the masked inputs for some comparison models that necessitate a consistent grid structure for both input and output data. 
As shown in Table~\ref{era5_generalization}, IPOT consistently outperforms all the baselines in all scenarios, demonstrating stable and strong performance. These results highlight the exceptional flexibility and accuracy of IPOT, showcasing its remarkable generalization capability.

\begin{table}[t]
\caption{Performance results for comparing the generalization ability on discretizations. The models are evaluated on the task of different resolutions and masked inputs for the ERA5 dataset.}
\label{era5_generalization}
\centering
\scriptsize
\begin{tabular}{c|ccc|c}
\specialrule{.15em}{.075em}{.075em}  
\multirow{ 2}{*}{Model} & \multicolumn{3}{c|}{Different resolutions} & Partial observed \\
\cline{2-5}
 & res$=4^{\circ}$ & res$=1^{\circ}$ & res$=0.25^{\circ}$ & Masked land\\
\specialrule{.15em}{.075em}{.075em}
FNO       & 1.30e--2 & 1.23e--2 & 1.24e--2 & 3.10e--2 \\ 
OFormer     & 3.66e--2 & 1.65e--2 & 1.86e--2 & 4.37e--2 \\ 
\hline
IPOT (ours) & \textbf{8.96e--3} & \textbf{7.78e--3} & \textbf{8.66e--3} & \textbf{2.83e--2} \\ 
\specialrule{.15em}{.075em}{.075em} 
\end{tabular}
\end{table}

\subsection{Ablation study}
\begin{figure}[t]
\centering
\includegraphics[scale=0.5]{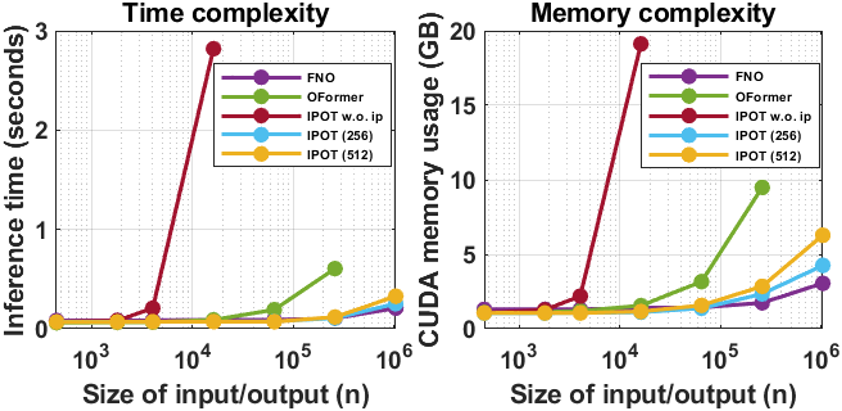}
\caption{Complexity comparisons on different resolutions. We compare the different models in terms of inference time (left) and CUDA memory usage (right) with different sizes of input/output.}
\label{complexity}
\vspace{-3mm}
\end{figure}

\textbf{Effect of the number of inducing points.}
We conducted experiments on ERA5 data with varying numbers of latent query vectors, ranging from 64 to 512, to evaluate the effect of the number of inducing points, and to emulate the quadratic complexity in attention blocks of the standard Transformer \cite{transformer}, we also consider IPOT without inducing points. In this variant, we use self-attention where the observational input is injected into query, key, and value components instead of using cross-attention with learnable queries in the encoder.
Table~\ref{number_ip} demonstrates that increasing the number of latent query vectors improves the performance of IPOT. Notably, when the number is over 256, IPOT sufficiently outperforms other baselines. However, when the number of inducing points is too small, IPOT exhibits poor performance. Additionally, the quadratic complexity of attention in a standard Transformer hinders convergence and results in excessive computational time. 
\begin{table}[t]
\scriptsize
\caption{Performance of IPOT with varying the number of inducing points. IPOT w.o ip denotes that IPOT without inducing points which emulates the standard Transformer.}
\label{number_ip}
\centering
\begin{tabular}{c|cccccc}
\specialrule{.15em}{.075em}{.075em}  
Model & $n$ & $n_z$ & $L$ & Runtime & Error & Complexity\\
\specialrule{.15em}{.075em}{.075em}
OFormer & 16.2K & 16.2K & 19 & 71.18 & 1.15e--2 & $\mathcal{O}(Lnd^2)$\\
IPOT w.o ip & 16.2K & 16.2K & 28 & $\gg100$ & -- &$\mathcal{O}(Ln^2d)$\\
\hline
IPOT (64)              & 16.2K & 64 & 28 & 7.44  & 1.45e--2 & $\mathcal{O}(Ln_z^2d)$\\
IPOT (128)             & 16.2K & 128 & 28 & 7.61  & 1.30e--2 & $\mathcal{O}(Ln_z^2d)$\\
IPOT (256)             & 16.2K & 256 & 28 & 7.91  & 6.87e--3 & $\mathcal{O}(Ln_z^2d)$\\
IPOT (512)             & 16.2K & 512 & 28 & 9.83  & 6.44e--3 & $\mathcal{O}(Ln_z^2d)$\\
\specialrule{.15em}{.075em}{.075em} 
\end{tabular}
\vspace{-3mm}
\end{table}
\\
\textbf{Computational complexity on different resolutions.}
We compared the computational complexity of different models on the ERA5 data at different resolutions, as shown in Figure~\ref{complexity}. The time complexities were assessed by measuring the inference time in seconds for processing observational data during testing, and the memory complexities were evaluated by recording CUDA memory allocation. It is observed that IPOT without latent inducing points, which serves as a surrogate for a standard Transformer with quadratic complexity in self-attention, does not scale well in terms of both time and memory costs. 
IPOT with 256 or 512 inducing points scale effectively up to $n=10^6$, making them suitable for handling large-scale observational data due to their efficiency in both time and memory complexity. 

\section{Conclusions}
\label{conclusion}
In this work, we raise potential issues on existing operator learning models for solving PDEs in terms of flexibility in handling irregular and arbitrary input and output discretizations, as well as the computational complexity for complex and long-term forecasting real-world problems. To address these issues, we propose IPOT, an attention-based architecture capable of handling arbitrary input and output discretizations while maintaining feasible computational complexity. This is achieved by using a reduced number of inducing points in the latent space, effectively addressing the quadratic scaling issue in the attention mechanism. 
Our proposed model demonstrates superior performance on a wide range of benchmark problems, including PDEs solved on both regular and irregular grids. Furthermore, we show that IPOT outperforms other baseline models on real-world data with competitive efficiency. 
Moreover, IPOT exhibits strong generalization ability, providing consistent performance in challenging real-life scenarios, while other models often suffer from significant performance degradation or require additional pre-processing for compatibility with the data structure. 

\bibliography{aaai24}

\section{Additional related Works}
\textbf{Learning-based PDE solvers.}
Recently, modern deep-learning techniques have shown great promise in solving PDEs. One popular direction is the physics-informed neural networks (PINNs) \cite{pinn, piml}, which use implicit neural representations (INRs) \cite{siren} to solve PDEs given boundary conditions and collocation points constrained to known governing equations. While PINNs yield mesh-agnostic and high-fidelity solutions, they require full knowledge of the PDEs and cannot be reused for new boundary conditions.
Another research direction attempts to predict flow fields around various shapes using the PointNet \cite{pointnet} on complex irregular grids \cite{pn}, which was later extended to include physics constraints \cite{pipn}. 
Additionally, \cite{meshfreeflow, ganflow} encodes low-resolution inputs to latent contexts and feeds them to decoding networks to predict solutions at output queries. While these methods were originally formulated for super-resolution problems and use physics constraints, our proposed approach is designed for operator learning problems. 
Similar to our work, mesh-agnostic approaches without any physics constraints have been achieved by \cite{nif, dino} using INRs conditioned on hypernetworks \cite{hypernet}. They built their architecture upon INRS, whereas our approach is built upon Transformers.

\section{Derivation of equation 4}
\label{Details_explanation}
Here is a brief explanation of the approximation with the integral for the cross-attention mechanism, where the softmax for the attention matrix is ignored for simplicity. 
\begin{equation}
\begin{aligned}
\textsl{Attn}(Y, X, X) = \begin{bmatrix} 
\textsl{Attn}(y_1, X, X) \\ 
\vdots \\ 
\textsl{Attn}(y_{n_y}, X, X) \end{bmatrix} \\
= \begin{bmatrix}
q(y_1) \\
\vdots \\[15pt]
q(y_{n_y}) \end{bmatrix} \begin{bmatrix}
k_1(x_1) & \ldots & k_1(x_{n_x}) \\
\vdots & \ddots & \vdots \\[15pt]
k_{d_q}(x_1) & \ldots & k_{d_q} (x_{n_x}) 
\end{bmatrix} \begin{bmatrix}
v(x_1) \\
\vdots \\
v(x_{n_x})
\end{bmatrix} \\[15pt]
= \begin{bmatrix}
q(y_1) \cdot k(x_1) & \ldots & q(y_1) \cdot k(x_{n_x}) \\
\vdots & \ddots & \vdots \\
q(y_{n_y}) \cdot k(x_1) & \ldots & q(y_{n_y}) \cdot k(x_{n_x})
\end{bmatrix} \begin{bmatrix}
v(x_1) \\
\vdots \\
v(x_{n_x}) \end{bmatrix} \\[15pt]
= \begin{bmatrix}
\sum_{i=1}^{n_x} (q(y_1) \cdot k(x_i)) v(x_i) \\
\vdots \\
\sum_{i=1}^{n_x} (q(y_{n_y}) \cdot k(x_i)) v(x_i)
\end{bmatrix} \\[15pt]
\approx \begin{bmatrix}
\int_{\Omega_x} (q(y_1) \cdot k(x)) v(x) dx \\
\vdots \\
\int_{\Omega_x} (q(y_{n_y}) \cdot k(x)) v(x) dx
\end{bmatrix} = \int_{\Omega_x} (q(Y) \cdot k(x)) v(x) dx.
\end{aligned}
\end{equation}
Here, the discretization of input is $\{x_1, ..., x_{n_x} \}$ (as key and value vectors), and it can be changed to the discretization of output $\{y_1, ..., y_{n_y} \}$ (as query vectors) with cardinality changed from $n_x$ to $n_y$. Using this mechanism, we can detach the dependences of discretization formats of input and output from the processor, by encoding arbitrary discretization $\{x_1, ..., x_{n_x} \}$ to a fixed size ($n_z$) of learnable latent set vectors, and decoding the latent set vectors to output arbitrary discretization $\{y_1, ..., y_{n_y} \}$. The discretization number is varied as $n_x$ (arbitrary) $\rightarrow$ $n_z$ (fixed) $\rightarrow$ $n_y$ (arbitrary).

\section{Attention blocks}
\label{appendix_attention_modules}
Mesh-independent neural operator (IPOT) consists of two types of attention blocks, cross- and self-attention blocks, which implement the respective attention mechanisms. The attention blocks have the shared structures following the Transformer-style architectures \cite{transformer, perceiver, perceiver_io}, which takes two input arrays, a query input $Y \in \mathbb{R}^{n_y \times d_y}$ and a key-value input $X \in \mathbb{R}^{n_x \times d_x}$,
\begin{equation}
\begin{aligned}
\label{attentional_module}
O = Y + \textsl{Attn}(\textsl{LN}(Y), \textsl{LN}(X), \textsl{LN}(X)), \\
\textsl{Attention}(Y, X, X) = O + \textsl{FF}(\textsl{LM}(O)), \\
\end{aligned}
\end{equation}
where $\textsl{LN}$ is layer normalization \cite{layernorm}, \textsl{FF} consists of two point-wise feedforward neural networks with a GELU nonlinearity \cite{gelu}, and the exact calculation of attention \textsl{Attn} is
\begin{equation}
\begin{aligned}
\label{attention}
\textsl{Attn}(X^q, X^k, X^v) = \textsl{Softmax}\left( \frac{QK^T}{\sqrt{d_q}} \right) V,
\end{aligned}
\end{equation}
where $Q = X^q W^q \in \mathbb{R}^{n_y \times d_q}$, $K = X^k W^k \in \mathbb{R}^{n_x \times d_q}$, and $V = X^v W^v \in \mathbb{R}^{n_x \times d_v}$ for a single headed attention. In the case of multi-headed attention, several outputs from different learnable parameters are concatenated and projected with the linear transformation. 

\section{Datasets}
\label{benchmarks}
In this section, we present an overview of the datasets and describe the corresponding equations and tasks along with the details of inputs and outputs. Table~\ref{dataset_overview} presents a summary of the datasets for easy reference across different domains, including problems of PDEs solved on regular grids, irregular grids, and real-world data. The footnotes of each problem set are the corresponding URL addresses for the datasets, which are publicly available. The train/test split settings are also directly taken from respective benchmark datasets. 
\begin{table*}[t]
\scriptsize
\caption{Overview of datasets and the corresponding tasks in terms of size of input and output.}
\label{dataset_overview}
\centering
\begin{tabular}{c|ccccc}
\specialrule{.15em}{.075em}{.075em}  
Dataset & Problem & Input & Output & Input size & Output size\\
\specialrule{.15em}{.075em}{.075em}
Burgers & 1D regular grid & $u(\cdot, 0)$ & $u(\cdot, 1)$ & 1,024$\times$1 & 1024$\times$1 \\
Darcy flow & 2D regular grid & $a$ & $u$ & 7,225$\times$1 & 7,225$\times$1 \\
Navier-Stokes, $\nu{=}$1e--3 & 2D regular grid & $w(\cdot, t)|_{t\in [0, 10]}$ & $w(\cdot, t)|_{t\in (10, 50]}$ & 4,096$\times$10 & 4,096$\times$1$\times$40 \\
Navier-Stokes, $\nu{=}$1e--4 & 2D regular grid & $w(\cdot, t)|_{t\in [0, 10]}$ & $w(\cdot, t)|_{t\in (10, 30]}$ & 4,096$\times$10 & 4,096$\times$1$\times$20 \\
Navier-Stokes, $\nu{=}$1e--5 & 2D regular grid & $w(\cdot, t)|_{t\in [0, 10]}$ & $w(\cdot, t)|_{t\in (10, 20]}$ & 4,096$\times$10 & 4,096$\times$1$\times$10\\
\specialrule{.15em}{.075em}{.075em}
Airfoil & Transonic flow & Mesh point & Velocity & 11,271$\times$2 & 11,271$\times$1 \\
Elasticity & Hyper-elastic material & Point cloud & Stress & 972$\times$2 & 972$\times$1 \\
Plasticity & Plastic forging & Boundary condition & Displacement & 62,620$\times$1 & 62,620$\times$4  \\
Spherical shallow water & Spherical manifold & $u(\cdot, 0)$ & $u(\cdot, t)|_{t\in (0, 20]}$ & 8,192$\times$2 & 8,192$\times$2$\times$19 \\
\specialrule{.15em}{.075em}{.075em}
ERA5 & [Train] $T_{2m}$ forecasting & $u(\cdot, t)|_{t \in [-6, 0]}$ & $u(\cdot, t)|_{t \in [1, 7]}$ & 16,200$\times$7 & 16,200$\times$1$\times$7 \\
ERA5 & [Test] Masked land & $u(\cdot, t)|_{t \in [-6, 0]}$ & $u(\cdot, t)|_{t \in [1, 7]}$ & 11,105$\times$7 & 16,200$\times$1$\times$7 \\
ERA5 & [Test] Masked sea & $u(\cdot, t)|_{t \in [-6, 0]}$ & $u(\cdot, t)|_{t \in [1, 7]}$ & 5,095$\times$7 & 16,200$\times$1$\times$7 \\
\specialrule{.15em}{.075em}{.075em}
\end{tabular}
\end{table*}

\subsection{Problems of PDEs solved on regular grids}
\textbf{Burgers' equation\footnote{\label{regular}\url{https://github.com/neuraloperator/neuraloperator}}.}
Burgers' equation is a non-linear parabolic PDE combining the terms of convection and diffusion. The benchmark problem of 1D Burgers' equation with periodic boundary conditions is defined as
\begin{gather*}
\label{burgers_eq}
\partial_t u(x, t) + \partial_x (u^2(x, t) / 2) = \nu \partial_{xx} u(x, t), \\
\qquad\qquad\!\! \ u(x, 0) = u_0 (x),  
\end{gather*}
where $u_0 \sim \mu$ is the initial state generated from $\mu = \mathcal{N}(0, 625(-\Delta+25I)^{-2})$ and $\nu = 0.1$ is the viscosity coefficient. The goal of operator learning is to learn mapping the initial state to the solution at $t=1$, $\mathcal{G}: u(\cdot, 0) \mapsto u(\cdot, 1)$. During training, the input and output discretizations are given at 1,024 equispaced grids. 
\\
\textbf{Darcy flow\footref{regular}.}
Darcy flow is a second-order elliptic PDE describing the flow of fluid through a porous medium. The benchmark problem of 2D steady-state Darcy flow on unit cell is defined as
\begin{gather*}
\label{darcy_flow}
-\nabla \cdot (a(x)\nabla u(x)) = f(x), \\ 
u(x) = 0, 
\end{gather*}
where $u$ is density of the fluid, $a \sim \mu$ is the diffusion field generated from $\mu = \mathcal{N}(0, (-\Delta + 9I)^{-2})$ with fixed forcing function $f = 1$. The goal of operator learning is to learn mapping the diffusion field to the solution of the density, $\mathcal{G}: a \mapsto u$. During training, the discretizations of input and output are sampled at 85${\times}$85${=}$7,225 regular grids. 
\\
\textbf{Navier-Stokes equation\footref{regular}.}
Navier-Stokes equation describes the dynamics of a viscous, incompressible fluid. The benchmark problem of the 2D Navier-Stokes equation in vorticity form on the unit torus is defined as 
\begin{gather*}
\label{ns_eq}
\ \partial_t w(x, t) + u(x, t) \cdot \nabla w(x, t) = \nu \Delta w(x, t) + f(x), \\ 
\nabla \cdot u(x, t) = 0, \\ 
w(x, 0) = w_0 (x),  
\end{gather*}
where $u$ is the velocity field, $w = \nabla \times u$ is the vorticity field, $w_0 \sim \mu$ denotes the initial vorticity field generated from $\mu = \mathcal{N}(0, 7^{3/2} (-\Delta + 49I)^{-2.5})$ with periodic boundary conditions, and the forcing function is kept $f(x) = 0.1 \left( \sin \left(2 \pi \left(x_1 + x_2\right)\right) + \cos \left(2 \pi \left(x_1 + x_2\right)\right)\right)$. 
The goal of operator learning is to learn the mapping from the initial times $t\in [0, 10]$ of the vorticity fields to further trajectories up to $t=T$, defined by $w(\cdot, t)|_{t \in [0, 10]} \mapsto w(\cdot, t)|_{t \in (10, T]}$.
The time total duration of each trajectory is $T{=}$50, 30, and 10 corresponding to the viscosity coefficients $\nu{=}$1e--3, 1e--4, and 1e--5, respectively, for each dataset \cite{fno}. 1,000 instances are used for training and 100 for testing. 
During both training and testing, the spatial discretizations of input and output are sampled at 65${\times}$65${=}$4,096 regular grids.

\subsection{Problems of PDEs solved on irregular grids}
\textbf{Airfoil\footnote{\label{irregular}\url{https://github.com/neuraloperator/Geo-FNO}}.}
The benchmark problem of the airfoil is based on the transonic flow over an airfoil, which is governed by Euler's equation. The equation is defined as
\begin{equation}
\begin{aligned}
\label{airfoil_eq}
\partial_t \rho + \nabla \cdot (\rho^f v) = 0, \\
\partial_t (\rho u) + \nabla \cdot (\rho u \otimes u + p I) = 0, \\
\partial_t E + \nabla \cdot \big( (E + p) u \big) = 0,
\end{aligned}
\end{equation}
where $\rho$ is the density, $u$ is the velocity field, $p$ is the pressure, and $E$ is the total energy. The far-field boundary conditions are $\rho_{\infty} = 1, p_{\infty} = 1$, and Mach number $M_{\infty}=0.8$, with no penetration imposed at the airfoil. 
The shape and mesh grids of the datasets used in this study are adopted from \cite{geo_fno}, which consists of variations of the NACA-0012 airfoils. The dataset is split into 1,000 instances for training and 200 instances for testing. The inputs and outputs are given as the mesh grids and the corresponding velocities in terms of Mach number.
\\
\textbf{Elasticity\footref{irregular}.}
The benchmark problem of elasticity involves hyper-elastic material, which is governed by constitutive equations. The equation on the bounded unit cell $\Omega\in[0, 1]^2$ is defined as 
\begin{equation}
\begin{aligned}
\label{elasticity_eq}
\rho \frac{\partial^2 u}{\partial t^2} + \nabla \cdot \sigma = 0,
\end{aligned}
\end{equation}
where $\rho$ is mass density, $u$ is the displacement vector, and $\sigma$ is the stress tensor, respectively. The random-shaped cavity at the center has a radius that is sampled as $r=0.2+\frac{0.2}{1+\exp(\tilde{r})}$, where $\tilde{r}$ is drawn from a distribution $\tilde{r}\sim\mathcal{N}(0, 4^2(-\nabla + 3^2)^{-1})$. 
The unit cell is fixed on the bottom edges and a tension traction of $t=[0, 100]$ is applied on the top edge. 
The hyper-elastic material used is the incompressible Rivlin-Saunders material. 
The dataset and split setting are also directly taken from \cite{geo_fno}, where 1,000 instances are for training and 200 for testing. The inputs and outputs are given as coordinates of point clouds and the corresponding stress. 
\\
\textbf{Plasticity\footref{irregular}.}
The plasticity benchmark problem involves 3D plastic forging, where the governing constitutive equation is the same as Equation~\ref{elasticity_eq} over time. The objective is to learn a mapping from the initial boundary condition to the mesh grids and displacement vectors over time. The target solution has the dimension of 62,620${\times}$4, where the output queries are structured on a 101${\times}$31 mesh grid over 20 time steps. This results in a total of 62,620 output queries, each consisting of 2 mesh grids and 2 displacement vectors, thus yielding 4 dimensions. The dataset and split setting are also directly taken from \cite{geo_fno}, where 900 instances are for training and 80 for testing. 
\\
\textbf{Spherical shallow water\footnote{\label{dino}\url{https://github.com/mkirchmeyer/DINo}}.}
We consider another PDE problem of the 3D spherical shallow water equation \cite{shallow, dino}. The shallow water equation is
\begin{equation}
\label{sha_eq}
\partial_t u = -f k \times u - g \nabla h + \nu \Delta u, \quad \partial_t h = -h \nabla \cdot u + \nu \Delta h,
\end{equation}
where $k$ is the unit normal vector to the spherical surface, $u$ is the velocity field tangent to the spherical surface, $w = \nabla \times u$ is the vorticity field, and $h$ is the thickness of the sphere. The observational data at time $t$ is given as $v_t = (w_t, h_t)$. $f, g, \nu, \Omega$ are the parameters of the Earth, which can be found in \cite{shallow}. We follow \cite{dino} to create symmetric phenomena on the northern and southern hemispheres for the initial conditions. $u_0(\phi, \theta)$ is the initial zonal velocity written as:
\begin{equation*}
u_0(\phi, \theta) = \begin{cases}
\left(\frac{u_{max}}{e_n} e^{ \frac{1}{(\phi - \phi_0)(\phi - \phi_1)} }, 0 \right) &\text{if $\phi \in (\phi_0, \phi_1)$}, \\
\left( \frac{u_{max}}{e_n} e^{ \frac{1}{(\phi + \phi_0)(\phi + \phi_1)} }, 0 \right) &\text{if $\phi \in (-\phi_1, -\phi_0)$}, \\
(0, 0) &\text{otherwise}.
\end{cases}
\end{equation*}
where latitude and longitude $\phi, \theta \in \left[-\frac{\pi}{2}, \frac{\pi}{2}\right] \times [-\pi,\pi]$, $u_{max} \sim \mathcal{U}(60, 80)$ is the maximum velocity sampled from uniform distribution range from 60 to 80, $\phi_0 = \frac{\pi}{7}$, $\phi_1 = \frac{\pi}{2} - \phi_0$, and $e_n = \exp{\left( -\frac{4}{(\phi_1-\phi_0)^2}\right)}$. We also follow \cite{dino} to initialize the perturbed water height $h_0(\phi, \theta)$ from the original one:
\begin{equation*}
h_0(\phi, \theta) = \hat{h} \cos{\phi} e^{-\left(\frac{\theta}{\alpha}\right)^2 } \left[ e^{ -\left( \frac{\phi_2 - \phi}{\beta}\right)^2 } + e^{-\left( \frac{\phi_2 + \phi}{\beta}\right)^2} \right]
\end{equation*}
where $\phi_2 = \frac{\pi}{4}$, $\hat{h} = 120 m$, $\alpha=\frac{1}{3}$, $\beta = \frac{1}{15}$ are constants from \cite{shallow}.

\subsection{Problem of real-world data}
\textbf{ERA5 reanalysis\footnote{\url{https://www.ecmwf.int/en/forecasts/datasets/browse-reanalysis-datasets}}.}
The ERA5 reanalysis database, provided by the European Centre for Medium-Range Weather Forecasts (ECMWF) \cite{era5}, offers extensive hourly and monthly measurements for various parameters, including temperature, precipitation, and wind speed at different pressure levels. 
For our experiments, we downloaded a subset of the ERA5 database, specifically for the 2m temperatures at 00:00 UTC from January 1st, 2018 to March 31st, 2023. 
We divided our dataset into non-overlapping segments of 7 consecutive days. The goal is to map the temperature field of the previous 7 days to the temperature field of the next 7 days, $u(\cdot, t)|_{t\in [-6, 0]} \mapsto u(\cdot, t)|_{t \in [1, 7]}$. The dataset comprised a total of 275 input-output pairs, where 250 instances were used for training and 25 instances were used for testing. 
We downsampled each frame by a factor of 8, resulting in a resolution of 2$^\circ$ (90$\times$180) for training, and we tested the pre-trained models on various resolutions, ranging from 0.25$^\circ$ to 4$^\circ$, as well as on masked input scenarios. 
Masked input with a land mask leads to 11,105 points, while the sea region contains 5,095 points. 

\section{Implementation details}
\label{appendix_implementation_details}
\begin{table*}[htbp]
\centering
\caption{Architecture details of IPOT.}
\label{implementation_table_architecture}
\scriptsize
\begin{tabular}{c|ccc|cccc|c}
\specialrule{.15em}{.075em}{.075em} 
Dataset & \multicolumn{3}{c|}{Regular} & \multicolumn{4}{c|}{Irregular} & Real \\
\hline
Problem & Burgers & Darcy flow & Navier-Stokes & 
Airfoil & Elasticity & Plasticity & Shallow water & ERA5 \\
\specialrule{.15em}{.075em}{.075em}
\multicolumn{9}{c}{Positional encoding} \\
\specialrule{.15em}{.075em}{.075em} 
Frequency bins  & 64 & [32, 32] & [12, 12] &
[8, 8] & [16, 16] & [3, 3, 3] & [20, 20, 20] & [64, 64]\\
Max frequency & 64 & [32, 32] & [20, 20] &
[16, 16] & [16, 16] & [12, 12, 12] & [32, 32, 32] & [64, 128]\\
Positional encoding & 1,024${\times}$129 & 7,225${\times}$130 & 4,096${\times}$50 &
11,271${\times}$34 & 972${\times}$132 & 62,620${\times}$21 & 8,192${\times}$123 & 16,200${\times}$258\\ 
\specialrule{.15em}{.075em}{.075em} 
\multicolumn{9}{c}{Encoder} \\
\specialrule{.15em}{.075em}{.075em} 
Input function values
& 1,024${\times}$1 & 7,225${\times}$1 & 4,096${\times}$1 &
11,271${\times}$1 & 972${\times}$1 & 62,620${\times}$1 & 8,192${\times}$1 & 16,200${\times}$1\\
Latent channels      & 64 & 64 & 128 &
64 & 64 & 64 & 128 & 128 \\
Number of heads      & 8  & 1  & 1 &
1 & 1 & 1 & 2 & 1\\
Inputs
& 1,024${\times}$130 & 7,225${\times}$131 & 4,096${\times}$51 &
11,271${\times}$35 & 972${\times}$133 & 62,620${\times}$22 & 8,192${\times}$124 & 16,200${\times}$259\\
\specialrule{.15em}{.075em}{.075em} 
\multicolumn{9}{c}{Processor} \\
\specialrule{.15em}{.075em}{.075em} 
Learnable queries
& 256${\times}$64 & 256${\times}$64 & 512${\times}$128 &
128${\times}$64 & 512${\times}$64 & 256${\times}$64 & 256${\times}$128 & 512${\times}$128\\
Latent channels          & 64 & 64 & 128 &
64 & 64 & 64 & 128 & 128\\
Number of heads          & 8 & 8 & 4 &
4 & 4 & 2 & 8 & 8\\
Number of blocks       & 1 & 4 & 2 &
2 & 4 & 2 & 2 & 4\\  
\specialrule{.15em}{.075em}{.075em} 
\multicolumn{9}{c}{Decoder} \\
\specialrule{.15em}{.075em}{.075em} 
Output queries
& 1,024${\times}$1 & 7,225${\times}$2 & 4,096${\times}$2 &
11,271${\times}$2 & 972${\times}$4 & 62,620${\times}$3 & 8,192${\times}$3 & 16,200${\times}$2\\
Latent channels         & 64 & 64 & 128 &
64 & 64 & 64 & 128 & 128\\
Number of heads      & 8  & 1  & 1 &
1 & 1 & 1 & 2 & 1\\
Outputs
& 1,024${\times}$1 & 7,225${\times}$1 & 4,096${\times}$1 &
11,271${\times}$1 & 972${\times}$1 & 62,620${\times}$4 & 8,192${\times}$2 & 16,200${\times}$1\\
\specialrule{.15em}{.075em}{.075em} 
\end{tabular}
\end{table*}

\begin{table*}[htpb]
\centering
\caption{Training details for IPOT.}
\label{implementation_table_training}
\scriptsize
\begin{tabular}{c|ccc|cccc|c}
\specialrule{.15em}{.075em}{.075em} 
Data type & \multicolumn{3}{c|}{Regular} & \multicolumn{4}{c|}{Irregular} & Real \\
\hline
Problem & Burgers & Darcy flow & Navier-Stokes & 
Airfoil & Elasticity & Plasticity & Shallow water & ERA5 \\
\specialrule{.15em}{.075em}{.075em}
Batch size & 20 & 10 & 100 & 20 & 10 & 10 & 64 & 10\\
Epochs & 2,000 & 1,000 & 5,000 & 1,600 & 1,600 & 1,600 & 3,000 & 5,000\\
Learning rate & 0.001 & 0.001 & 0.001 & 0.001 & 0.001 & 0.001 & 0.001 & 0.001 \\
Learning rate decay & 0.5 & 0.5 & 0.5 & 0.5 & 0.5 & 0.5 & 0.5 & 0.5 \\
Step decay & 250 & 200 & 250 & 200 & 200 & 200 & 500 & 500\\
\specialrule{.15em}{.075em}{.075em} 
\end{tabular}
\end{table*}

\subsection{Training.} 
The experiments are conducted on a 24GB NVIDIA GeForce RTX 3090 GPU and use AdamW optimizer \cite{adamw} with an initial learning rate of 1e--3. The implemented architectures and corresponding training hyperparameters for each problem are summarized in 
Table~\ref{implementation_table_architecture} and Table~\ref{implementation_table_training}, respectively. 

\subsection{Baselines}
\begin{figure}[t]
\centering
\vspace{-3mm}
\includegraphics[scale=0.65]{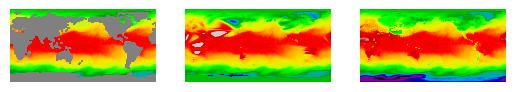}
\caption{Land regions are masked which is motivated by situations when observations are only available for the sea surface (left). In order to adopt FNO to masked input tasks, we incorporated interpolated values (middle). However, there can be a large gap between the real temperature field (right) and the interpolated inputs.}
\label{interpolated_inputs}
\vspace{-3mm}
\end{figure}

The results for the problems on regular grids and irregular grids were obtained from the related literature, including DeepONet \cite{deeponet1}, the mesh-based learner with graph neural networks (Meshgraphnet) \cite{epd2}, Fourier neural operator (FNO) \cite{fno}, Factorized-Fourier neural operator (FFNO) \cite{ffno}, and operator Transformer (OFormer) \cite{oformer}.
For the evaluation of baselines on real data (ERA5), we produced the results using their original codes of FNO\footnote{\url{https://github.com/neuraloperator/neuraloperator}} and OFormer\footnote{\url{https://github.com/BaratiLab/OFormer}}.
Since the problem is formulated as a time-stepping system similar to the Navier-Stokes equation, we adopted almost the same architectures that were used in the Navier-Stokes problem for the temperature forecasting task. 
However, due to the requirement of having the input and output
share the same regular grid structure in FNO \cite{operator_overview, fno}, incorporating interpolated values on the masked region is necessary for masked input tasks. To adopt FNO to masked inputs, we use cubic interpolation methods to obtain interpolated input values for land or sea coordinates. These interpolated values are then combined with the masked inputs, as shown in Figure~\ref{interpolated_inputs}. However, due to the masked regions being irregular and complex, the interpolated input values may not accurately represent the real temperature fields. As a result, this leads to significant performance degradation for FNO in forecasting the output temperature fields. On the other hand, IPOT does not require such problem-specific pre-processing. 

\section{Additional results}
\label{appendix_additional_results}
\subsection{Long-term predictions}
We conducted experiments with extended long-term predictions with IPOT and DINO \cite{dino} (state-of-the-art comparison) in Table~\ref{long_term}. The original test set from \cite{dino} consists of long trajectories (160 frames) each of which is divided into 8 short trajectories (20 frames). We modified the slicing scheme so that each long trajectory is divided into 4 short trajectories (40 frames). Table~\ref{long_term} includes extended predictions on both approaches on $t=20-
30$ and $t=30-40$. The result supports our model's stability in producing accurate long-term predictions. Also, we provide the visualization of predictions of IPOT on $t=0, 10, 20, 30$, and $40$ in Figure 4 (left) in the main text.
\begin{table}[htpb]
\vspace{-3mm}
\hspace{2em}
\scriptsize
\centering
\caption{Long-term predictions of DINO and IPOT on spherical shallow water. }
\label{comparison_table}
\begin{tabular}{c|cccc}
\specialrule{.15em}{.075em}{.075em}
\multirow{ 2}{*}{Model} & \multicolumn{4}{c}{Shallow water} \\
\cline{2-5}
& $t$=0-10 & $t$=10-20 & $t$=20-30 & $t$=30-40 \\
\specialrule{.15em}{.075em}{.075em} 
DINO & \textbf{1.06e--4} & 6.47e--4 & 1.37e--3 & 1.52e--03 \\
IPOT & 1.74e--4 & \textbf{5.12e--4} & \textbf{1.00e--3} & \textbf{1.11e--03} \\
\specialrule{.15em}{.075em}{.075em} 
\end{tabular}
\label{long_term}
\end{table}

\subsection{Compared with Perceiver IO}
We conducted experiments with long-term predictions with IPOT and perceiver io \cite{perceiver_io} in Table~\ref{comparison_table_v}. 
\begin{table}[htpb]
\centering
\scriptsize
\caption{Performance comparisons with Perceiver IO on Navier-Stokes equations.}
\label{comparison_table_v}
\begin{tabular}{c|ccc}
\specialrule{.15em}{.075em}{.075em}
\multirow{ 2}{*}{Model} & \multicolumn{3}{c}{Navier Stokes} \\
\cline{2-4}
 & $\nu$=1e--3 & $\nu$=1e--4 & $\nu$=1e--5 \\
\specialrule{.15em}{.075em}{.075em} 
Perceiver IO & 1.83e--2 & 2.59e--1 & 2.76e--1 \\
IPOT & \textbf{8.85e--3} & \textbf{1.22e--1} & \textbf{1.48e--1} \\
\specialrule{.15em}{.075em}{.075em} 
\end{tabular}
\end{table}

\end{document}